\documentclass[preprint,12pt,times,authoryear]{elsarticle}

\usepackage{amssymb}
\usepackage{amsmath}

\usepackage{xurl}
\usepackage[hidelinks]{hyperref}
\journal{}

\begin{document}

\begin{frontmatter}



\title{Distinct dynamics of conceptual and referential disruptions in human reading and large language model processing} 

\author[upf]{Rui He}
\author[upf]{Nihal Altay}
\author[upf,icrea]{Wolfram Hinzen}

\affiliation[upf]{organization={Grammar and Cognition Lab, Department of Translation \& Language Sciences, Universitat Pompeu Fabra},
                  city={Barcelona},
                  country={Spain}}

\affiliation[icrea]{organization={Institut Catal\`a de Recerca i Estudis Avan\c{c}ats (ICREA)},
                    city={Barcelona},
                    country={Spain}}

\begin{abstract}

Linguistic meaning is grounded in conceptual content, from which reference to particular entities emerges as words enter discourse. To examine the processing dynamics associated with these two dimensions of meaning, we selectively disrupted conceptual or referential information in short narratives and traced the resulting effects in human self-paced reading and in the predictive and representational processing of large language models. In human reading, conceptual disruptions produced a strong but localized processing cost, emerging immediately after the distorted word, reaching an early maximum, and then declining rapidly. Referential disruptions produced weaker effects, which decreased more gradually across subsequent words, and were more strongly modulated by sentence boundaries. In the language model, both disruptions emerged immediately at the manipulated word. Contextual model surprisal showed a pattern closely paralleling human reading: conceptual disruption produced a larger, more locally concentrated effect that decayed rapidly, whereas referential disruption produced a smaller and more gradual downstream effect. 
Output-layer representations showed a different pattern: referential disruption produced a larger initial displacement, while both distortions were subsequently characterized by power-law decay. Together, these results provide convergent evidence for distinguishable processing dynamics of two types of meaning: conceptual information imposes a more locally concentrated integration cost, whereas referential information engages a more distributed process of maintaining discourse-level identity.
\end{abstract}

\begin{keyword}

conceptual meaning \sep referential meaning \sep self-paced reading \sep sentence boundary \sep surprisal \sep representational dynamics \sep large language models

\end{keyword}

\end{frontmatter}

\section{Introduction}
\label{sec1}
Language unfolds one word at a time, yet its interpretation must remain coherent across much longer stretches of sentences and discourse. Each incoming expression is interpreted in relation to what has already been established and, in turn, contributes to the linguistic context in which subsequent material is understood. Language thus carries its own interpretive history forward as the discourse continues, with prior linguistic material constraining later interpretation at multiple levels \citep{brilmayer_referential_2021, brothers_multiple_2023, dijksterhuis_pronouns_2024}. These influences can extend over different temporal spans, bearing most strongly on the immediate interpretation of an expression in some cases while remaining relevant farther downstream in others \citep{chen_cortical_2024}. How such influences propagate over time may therefore illuminate the linguistic relations through which interpretation is maintained.

Maintaining a linguistic interpretation requires more than the accumulation of conceptual content. Words contribute lexical-conceptual material whose interpretation is further shaped by context \citep{leclercq_redefining_2023}, but this content alone does not determine what an expression refers to on a particular occasion. This distinction has long been recognized in semantic theory. \cite{frege_uber_1892} distinguished between the way an expression presents its meaning and what it refers to, and later formal approaches examined how word meanings combine within sentences and how referents are tracked across discourse \citep{heim_file_2002, kamp_theory_2013, montague_universal_1970}. The grammatical account adopted here goes a step further by treating reference itself as partly determined by linguistic structure: lexical material is not inherently referential, but becomes referential through the grammatical configurations in which it occurs \citep{hinzen_grammar_2016}. Conceptual and referential meaning are thus closely coordinated in ordinary language without being reducible to the same form of linguistic organization. Despite this coordination, the distinction motivates two experimental perturbations that place differential pressure on the two. Conceptual distortion (CD) alters the lexical-conceptual content contributed by an expression so that it becomes incompatible with its established context, whereas referential distortion (RD) alters a linguistic relation required to maintain or recover the intended discourse referent. These manipulations provide differential perturbations of conceptual and referential organization while preserving their interdependence within the broader linguistic system.

As linguistic interpretation proceeds incrementally, the consequences of these perturbations may extend beyond the expression at which they are introduced, shaping the conditions under which subsequent material is interpreted. Theoretically relevant information may therefore lie not only in the magnitude of the processing difficulty they induce, but also in how this difficulty unfolds through the material that follows. Psycholinguistic research has extensively examined lexical-semantic and referential processing \citep{dietrich_discourse_2024, kutas_thirty_2011}, although these have more often been studied in separate paradigms than compared within a common perturbational framework. Self-paced reading (SPR) is well suited to such a comparison, because it traces behavioral consequences word by word using reading time (RT), and RT effects are known to extend beyond their lexical source into subsequent positions \citep{smith_effect_2013}. Recent evidence comes particularly close to the present concept-reference distinction. \cite{soberats_referential_2025} found sharply local effects for lexical-semantic anomalies but substantially more extended effects for grammatical anomalies affecting referential meaning. Although these manipulations do not map directly onto the present CD and RD conditions, this contrast raises the possibility that conceptual and referential perturbations differ not only in their immediate impact, but also in their downstream dynamics. In addition, such propagation need not depend on distance alone. If referential meaning is partly sustained by grammatical organization, then the persistence of referential perturbation may also be sensitive to the structural points at which an unfolding interpretation is completed or reorganized \citep{martin_grammar_2014}. Sentence boundaries are particularly relevant in this respect. They mark the completion of one configuration of linguistic structure before another begins, and can therefore provide a structural reset point for ongoing interpretation \citep{arsenijevic_absence_2012}. Sentence-final positions are also associated with increased integration or wrap-up during processing \citep{meister_analyzing_2022}. A perturbation that reaches a sentence boundary can therefore reveal whether its persistence follows linear distance alone or is reshaped by the grammatical structure through which discourse unfolds. Accordingly, rather than treating downstream effects only as a set of predefined spillover regions, we characterize their progression as a propagation profile, tracing how they emerge, persist, and dissipate across linguistic distance and structural boundaries.

This raises a parallel question for language models (LMs): if conceptual and referential perturbations leave different downstream signatures in human processing, are these differences also expressed in the way a computational language system carries linguistic context forward? LM hidden-state representations have been shown to predict neural responses during natural language processing, despite important differences between artificial and biological systems in how context is accumulated over time \citep{tikochinski_incremental_2025, yang_cross-lingual_2026}. In an autoregressive model, the influence of prior context is expressed both in the probabilities assigned to upcoming words and in the contextual representations from which these predictions arise. Contextual surprisal therefore provides a measure of how a perturbation reshapes subsequent lexical expectations, a quantity closely related to variation in human RT \citep{wilcox_testing_2023}. At the same time, beyond its immediate effect on prediction, the perturbation may remain detectable in the model’s evolving hidden states, which encode information from the surrounding linguistic context \citep{ethayarajh_how_2019}. \cite{skrill_large_2023} showed that such representational influence can persist across subsequent positions, decaying with distance while becoming increasingly sensitive to structural boundaries. Predictive and representational effects need not follow identical dynamics. A perturbation may remain detectable in the model’s evolving contextual state after its immediate effect on lexical expectation has diminished, or conversely, may strongly alter prediction while producing only a transient representational displacement. We therefore examine both contextual surprisal and output-layer representational distance, asking whether the distinct propagation profiles of conceptual and referential distortions observed in human reading are also expressed in model prediction, representation, or both.

Thus, we hypothesize that conceptual and referential perturbations differ not only in the magnitude of their effects, but in how those effects propagate through subsequent linguistic context. In human reading, conceptual distortion is expected to produce a more concentrated downstream effect, whereas referential distortion should remain consequential over a broader span and show greater sensitivity to sentence boundaries. To test these predictions, we designed a word-by-word self-paced reading experiment in which conceptual and referential distortions were introduced into extended discourse. We analyzed how their behavioral consequences propagated with linguistic distance and were modulated by sentence boundaries. We then applied the same perturbations to autoregressive language models, examining their effects separately in contextual surprisal and output-layer representations, with Qwen3-4B as the primary model and Llama-3.2-3B as a replication. This parallel design allows the propagation of conceptual and referential information to be compared across human behavior, model prediction, and model representation.

\section{Methods}\label{methods}
\subsection{Participants}\label{participants}
We conducted an online self-paced reading experiment with 99 participants recruited through Prolific. Participants were native speakers of English aged 20-50 years (mean age = 36.47 years; 60 females, 39 males). Ninety-five had completed high school education. The study followed ethical principles consistent with the guidelines of the CIREP (Institutional Committee for Ethical Review of Projects) at Universitat Pompeu Fabra. Formal ethics review was not required for this non-biomedical study under current regulations in Spain. A comprehension question followed each story. Participants were retained only if they achieved greater than 60\% accuracy across the comprehension questions. All participants in the final sample met this criterion, with a mean comprehension accuracy of 92.98\%. 

\subsection{Stimuli}\label{stimuli}
Participants read 21 fables adapted from a publicly available collection of Aesop’s fables (\url{https://aesopfables.com/aesopsel.html}) in a word-by-word, non-cumulative self-paced reading paradigm. The source texts were rewritten in modern English using ChatGPT (GPT-5.3-mini) with the prompt, \textit{“Can you rewrite this story in modern English?”} Stories were approximately 60–120 words long, with a median length of 95 words. Each story was adapted into three versions: Original, CD, and RD. The Original version served as the undistorted baseline. CD involved replacing a contextually appropriate noun with a semantically incongruent alternative (e.g., \textit{the rushing current → the rushing mountain}), creating a local lexical-semantic disruption. RD involved altering a pronoun or determiner so that it conflicted with the intended discourse referent (e.g., \textit{he discovered → she discovered}), disrupting reference tracking across the discourse. This distinction allowed us to compare the downstream consequences of conceptual and referential disruptions. Examples are provided in the supplementary. 

Distortions were inserted manually at positions that produced a clear disruption while preserving the broader narrative structure. Opening sentences were left undistorted to allow readers to establish the discourse context. To avoid confounding conceptual distortions with sentence-final wrap-up effects, an adjunct was added when a CD would otherwise occur sentence-finally (e.g., \textit{“He brought the donkey home and placed it in the stable with his other donkeys, so it could rest.”}). CD and RD occurred at approximate rates of one per 30 and one per 20 words, respectively.

A word-level stimulus template was generated by aligning each distorted sentence with its Original counterpart. The distorted words were labelled K; the three preceding and following positions were labelled K-3 to K-1 and K+1 to K+3; and the sentence-final word was additionally labelled K\_END. Positional alignment was used when sentence lengths were identical. Sequence alignment was used for the small number of sentences that differed in length.

The final stimulus set comprised 63 story versions. To assess their comparability, we calculated whole-story perplexity, mean lexical frequency, and mean word length. Perplexity was estimated with Qwen3-4B-Base. Lexical frequency was quantified using mean Zipf frequency from the Python package wordfreq. Word length was defined as the mean number of alphabetic characters per lexical token. Differences among Original, CD, and RD were evaluated across the 21 matched story sets using Friedman tests and Siegel–Castellan pairwise comparisons, with p values corrected following the false discovery rate (FDR) procedure. 

\subsection{Procedure}\label{procedure}
The experiment was administered online using Gorilla. Stories were presented word by word using a non-cumulative self-paced reading paradigm. Participants advanced to each successive word by pressing the spacebar. Before each story, a separate instruction screen prompted participants to press the spacebar to begin, providing a common starting point before word presentation. Each participant read 21 stories, but encountered only one version of each story: seven Original, seven CD, and seven RD stories. Stories were presented in randomized order. Each story was followed by a four-alternative comprehension question designed to assess attention and comprehension without explicitly drawing attention to the manipulated expressions.

\subsection{Data de-identification and
preprocessing}\label{data-de-identification-and-preprocessing}
The raw Gorilla dataset was de-identified before analysis. Original participant identifiers were replaced with sequential anonymous identifiers, and session identifiers were recoded within participant. Timestamps, completion codes, external session identifiers, device and browser information, monitor and viewport dimensions, and other unnecessary participant metadata were removed. Word-level RTs were reconstructed from the Gorilla event log. A word-presentation event was retained when it was immediately followed by the corresponding keyboard-response event, and RT was calculated as the difference between the two event timestamps. 

Observed word sequences were aligned with the corresponding stimulus template. Exact positional matching was used for trials with complete word sequences. Trials with up to three missing word events were retained only when a unique alignment with the stimulus sequence could be established. Trials containing unresolved token mismatches, ambiguous alignments, more than three missing words, or additional or duplicated word events were excluded. Following alignment and trial-level quality control, 1,972 of the 2,079 possible participant-story trials were retained (94.85\%): 664 Original, 657 CD, and 651 RD trials. Each story-condition combination therefore contributed data from approximately 31–32 participants on average, indicating broadly balanced coverage across conditions. Where more than one observation was available for the same participant and word position, observations were averaged to produce a single participant-level word observation.

\subsection{\texorpdfstring{Reading time preprocessing}{Reading time preprocessing }}\label{reading-time-preprocessing}
Reading times outside the range of 100--3000 ms were treated as outliers and replaced with the median RT for the corresponding participant and story \citep{lo_periodic_2023}. When a participant–story median was unavailable, the participant-level median was used instead. Imputation affected 0.32\% of data values (0.38\% in Original, 0.35\% in CD, and 0.22\% in RD). RTs were subsequently log-transformed to reduce positive skew. To account for systematic variation in reading speed unrelated to the experimental manipulation, log-transformed RTs were residualized prior to the main analyses. To account for spillover between successive self-paced reading responses, we followed the approach proposed by \cite{vasishth_proper_2006}, in which RT at the current word is adjusted for RT at the immediately preceding word. Log-transformed reading times were therefore residualized against lag-1 log RT using a mixed-effects model with a participant random intercept. Lag-1 reading time was defined only for contiguous words within the same participant, story, and story version. No lag was carried across missing word events. The resulting residualized log reading times were used as the dependent variable in subsequent analyses.

\subsection{Reading time analysis}\label{reading-time-analysis}
\paragraph{Local reading-time effects} Local distortion effects were examined from K-3 to K+3 and at K\_END. For each position, CD and RD were separately compared with their matched Original observations using mixed-effects linear models. Residualized log RT was the dependent variable, with condition as the predictor of interest and word position within the story, lexical surprisal, and word length as covariates. Models included random intercepts for participants and matched stimulus items. The condition coefficient represented the estimated distortion-related change in residualized log RT. $P$ values were FDR-adjusted across positions separately for CD and RD.

\paragraph{Downstream reading-time trajectories} We next characterized the persistence of distortion effects across subsequent words. For each distorted expression, the final K token was treated as the trajectory anchor, and distance was defined as the number of words following this position. Trajectories extended to the end of the sentence or to a maximum of K+10, whichever came first. Because the behavioral response emerged after presentation of the distorted word, the RT trajectory analysis covered K+1 to K+10. Separate linear mixed-effects models were fitted for CD and RD. Residualized log RT was modelled as a function of categorical word distance, condition, and their interaction. The models also included an interaction between condition and sentence-boundary status to test whether the distortion effect was additionally modulated at the final word of the sentence. Word position within the story, lexical surprisal, and word length were included as covariates, with participant-level random intercepts. A distance was retained only when both Original and distorted observations were available, with at least 20 observations and at least two matched items contributing data. These models therefore tested both whether the magnitude of the distortion effect changed across downstream positions and whether sentence-final words showed an additional distortion-related effect.

\paragraph{Functional form of the reading-time trajectories} We next examined the shape of the downstream RT trajectories. For each distortion type, the distance-specific effects estimated from the mixed-effects models were fitted with 11 candidate functions: constant, linear, quadratic, exponential, power-law, log-normal, log-Cauchy, Zipf–Alekseev, broken-stick, double-exponential, and exponential-plus-power. The signed distortion effects were retained so that both the magnitude and direction of the response were preserved. Curve fitting incorporated the covariance among the distance-specific estimates, thereby accounting for their statistical dependence. Multiple initial parameter values were used for nonlinear models to reduce sensitivity to model initialization. Candidate functions were compared using the corrected Akaike information criterion (AICc). The function with the lowest AICc was selected as the best-supported trajectory for each distortion type. Uncertainty in derived trajectory characteristics was quantified using 1,000 parametric bootstrap samples. For each bootstrap sample, a new trajectory was drawn from the multivariate distribution defined by the estimated distance effects and their covariance matrix, and the selected function was refitted. Percentile-based 95\% confidence intervals were then calculated for relevant trajectory characteristics, such as peak location and half-maximum range for peak-shaped trajectories and slope or overall change for monotonic trajectories. Detailed fitting procedures and function forms are described in the supplementary. 

\paragraph{Direct comparison of CD and RD reading-time trajectories} Because fitting CD and RD separately does not directly test whether their trajectories differ, we additionally compared the two distortion types within a single mixed-effects model. Residualized log RT was modelled as a function of categorical word distance, distortion type, and their interaction, together with the interaction between distortion type and sentence-boundary status. Word position, lexical surprisal, and word length were included as covariates, with participant-level random intercepts. The distance-by-distortion-type interaction tested whether the downstream profiles differed between CD and RD. We also compared an early window (K+1 to K+3) with a late window (K+8 to K+10) to assess relative persistence. This contrast tested whether the change from early to late positions differed between RD and CD.

\subsection{Large language model analysis}\label{large-language-model-analysis}
To examine whether the distinction observed in human reading behavior was also reflected in large language models (LLMs), we applied the same K-centered analyses and effect decay framework to two complementary LLM-derived measures: surprisal, indexing predictive processing, and representational distance, indexing changes in contextual representation.

The same stimuli and original-distorted word mappings were analyzed using the pretrained 4-billion-parameter Qwen3 base model (Qwen/Qwen3-4B-Base). The model was used in evaluation mode without fine-tuning, and each complete story version was processed in a single forward pass without added special tokens. Token-level surprisal was calculated from the model's autoregressive next-token probability distribution. When a word was split into multiple subword tokens, token-level surprisals were summed to obtain word-level surprisal. The distortion-related surprisal effect was defined as
\begin{equation*}
    \Delta S_{t} = S_{t}^{\text{distorted}} - S_{t}^{\text{Original}},
\end{equation*}
such that positive values indicated greater surprisal in the distorted sequence than at the corresponding position in the Original sequence.

To quantify representational change, we extracted the final-layer hidden state associated with the final subword token of each word. The divergence between matched distorted and original representations was quantified using cosine distance,
\begin{equation*}
    D_{t} = 1 - \cos( h_{t}^{\text{distorted}},h_{t}^{\text{Original}}).
\end{equation*}

The two measures capture complementary aspects of model processing: surprisal reflects the effect of a distortion on the model’s predictions, whereas cosine distance reflects the extent to which the distortion alters its contextual representation.

We then applied analyses parallel to those used for RT. First, local effects were examined at the same K-3 to K+3 and K\_END positions. For each measure, distortion type, and position, an intercept-only ordinary least-squares model was fitted to the matched distortion measure. Cluster-robust standard errors were calculated at the manipulated story–sentence level to account for non-independence among observations from the same distortion unit. Analyses were conducted separately for CD and RD and for surprisal and cosine distance. Positions with fewer than three observations or fewer than two independent distortion units were not modelled, and p values were adjusted within each measure and distortion type using FDR.

Second, we examined downstream trajectories using the same perturbation mappings as in the RT analysis. In contrast to RT, model trajectories included K itself and therefore extended from K to K+10, because both prediction and representation could respond directly at the distorted token. For each measure and distortion type, linear models were fitted with categorical word distance and sentence-boundary status as predictors. Cluster-robust standard errors were calculated at the perturbation-trajectory level, and only distances represented by at least 10 distinct trajectories were retained. The resulting trajectories were fitted with the same 11 candidate functional forms used for the RT analysis, using the same covariance-weighted fitting procedure. Models were compared using AICc. The model with the lowest AICc was treated as the best-supported functional form for each measure and distortion type. This common analysis framework allowed the downstream dynamics of conceptual and referential distortions to be compared across behavioral, predictive, and representational measures.

Finally, CD and RD were compared directly for each model-derived measure. Regression models included the interaction between categorical word distance and distortion type, together with the interaction between sentence-boundary status and distortion type, with trajectory-clustered inference. As in the RT analysis, a planned persistence contrast compared an early downstream window (K+1 to K+3) with a late window (K+8 to K+10) to test whether the early-to-late change differed between CD and RD.

To assess whether the observed model patterns generalized beyond a single architecture, the complete analysis was independently repeated using the pretrained 3-billion-parameter Llama-3.2 base model (meta-llama/Llama-3.2-3B), using the same stimuli, word mappings, model-derived measures, analysis windows, and statistical procedures.

\section{Results}\label{results}

\subsection{Stimulus characteristics}\label{stimulus-characteristics}

\begin{figure}[t]
	\centering
	\includegraphics[width=\textwidth]{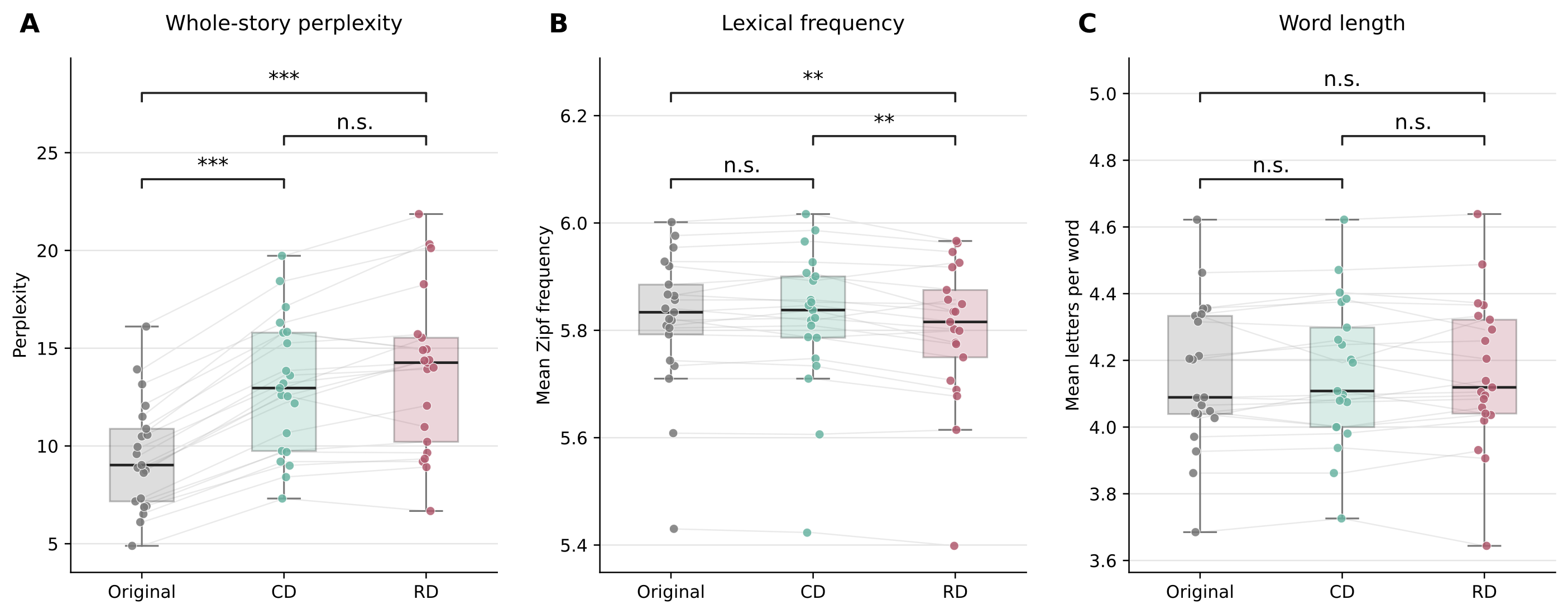}
	\caption{Comparability of stimulus versions on whole-story and lexical characteristics. Distributions of (A) whole-story perplexity, (B) mean lexical frequency (Zipf scale), and (C) mean word length across Original, conceptually distorted (CD), and referentially distorted (RD) versions of the 21 stories. Points represent individual stories, with grey connecting lines linking the three matched versions of each story. Pairwise differences were evaluated using Siegel–Castellan comparisons following the corresponding Friedman test. Significance labels indicate FDR-adjusted $p$ values: n.s., $p \geq 0.05$; *, $p < 0.05$; **, $p < 0.01$; ***, $p < 0.001$.}
	\label{FIG:1}
\end{figure}

The three stimulus versions were broadly comparable in surface characteristics, as shown in Figure~\ref{FIG:1}. Both CD and RD increased whole-story perplexity relative to the Original condition, as expected, whereas CD and RD did not differ significantly from each other, supporting their relative comparability in the overall magnitude of contextual disruption. RD showed a small difference in lexical frequency, likely reflecting the referential manipulation itself, as replacing pronouns or determiners can unavoidably alter lexical frequency because function words differ substantially in their baseline frequency (e.g., \textit{the vs. an}). Mean word length did not differ across conditions.

\begin{figure}[t]
	\centering
	\includegraphics[width=\textwidth]{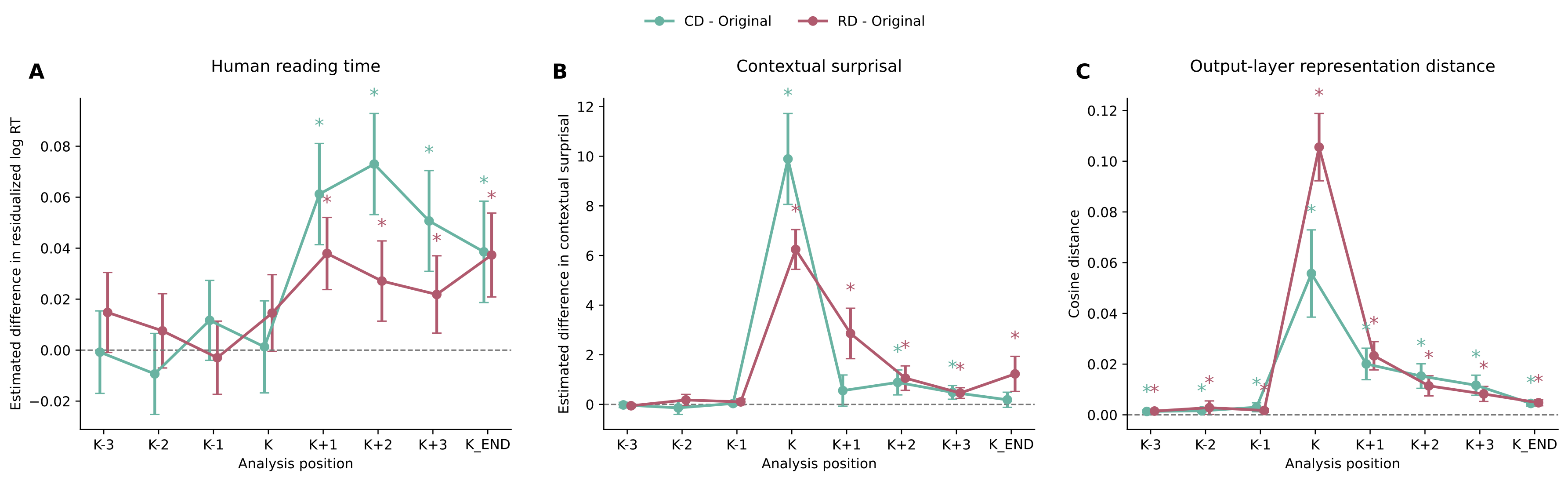}
	\caption{Local effects of conceptual and referential distortions in human reading and LLM processing. (A) Estimated effects of conceptual distortion (CD) and referential distortion (RD), relative to the matched Original condition, on residualized log reading time. (B) Changes in contextual surprisal produced by CD and RD in Qwen3-4B-Base, calculated relative to the corresponding Original words. (C) Output-layer representation distance between distorted and Original versions, quantified as cosine distance between the corresponding final-layer hidden states. K denotes the final distorted token; K-3 to K-1 and K+1 to K+3 denote positions relative to K, and K\_END denotes the sentence-final word. Points show estimated effects and error bars indicate 95\% confidence intervals. The horizontal dashed line indicates zero effect. Asterisks indicate effects that remained significant after FDR correction across analysis positions, applied separately for each distortion type within each outcome (FDR-adjusted 
    $p<0.05$).}
	\label{FIG:2}
\end{figure}

\subsection{Local effects of conceptual and referential distortions}\label{results_local}
Conceptual and referential distortions showed distinct temporal profiles in human reading behavior and in the language model (Figure~\ref{FIG:2}). In human RTs, neither distortion produced a significant effect at K; instead, effects emerged from K+1 onward (Figure~\ref{FIG:2}A). These downstream effects were observed in residualized reading times after accounting for the contribution of the immediately preceding word’s reading time, following the spillover-control approach proposed by \citep{vasishth_proper_2006}. They therefore cannot be attributed simply to the expected word-to-word carryover in response latency. CD produced a pronounced increase beginning at K+1, reaching its largest estimated effect around K+2 before decreasing toward the end of the sentence. RD produced a smaller post-perturbation increase, with effects distributed more evenly across subsequent positions and extending also to sentence final. Thus, both manipulations elicited delayed behavioral consequences, with CD characterized by a sharper post-perturbation increase. The LLM responded immediately at the distorted word. Contextual surprisal increased sharply at K for both distortions, with a larger peak for CD (Figure~\ref{FIG:2}B). The CD effect declined rapidly thereafter, whereas RD showed a smaller but more distributed downstream response, including K+1 and the sentence-final position. Output-layer representations showed a similarly immediate perturbation (Figure~\ref{FIG:2}C). Representation distance was maximal at K for both distortion types and declined substantially thereafter. Small pre-K distances were also significantly greater than zero, but these should not be interpreted as anticipatory effects, because the distorted and Original sequences could already differ owing to earlier perturbations within the same story and cosine distance is bounded at zero. Human and model responses therefore differed primarily in onset. LLM prediction and representation changed immediately at K with quick decay, whereas the behavioral cost emerged from K+1 onward. Both nevertheless distinguished the downstream consequences of conceptual and referential disruption, motivating the trajectory analyses below. 

\subsection{Temporal dynamics of conceptual and referential distortion effects in reading time}\label{results_RT}
\begin{figure}
	\centering
	\includegraphics[width=0.87\textwidth]{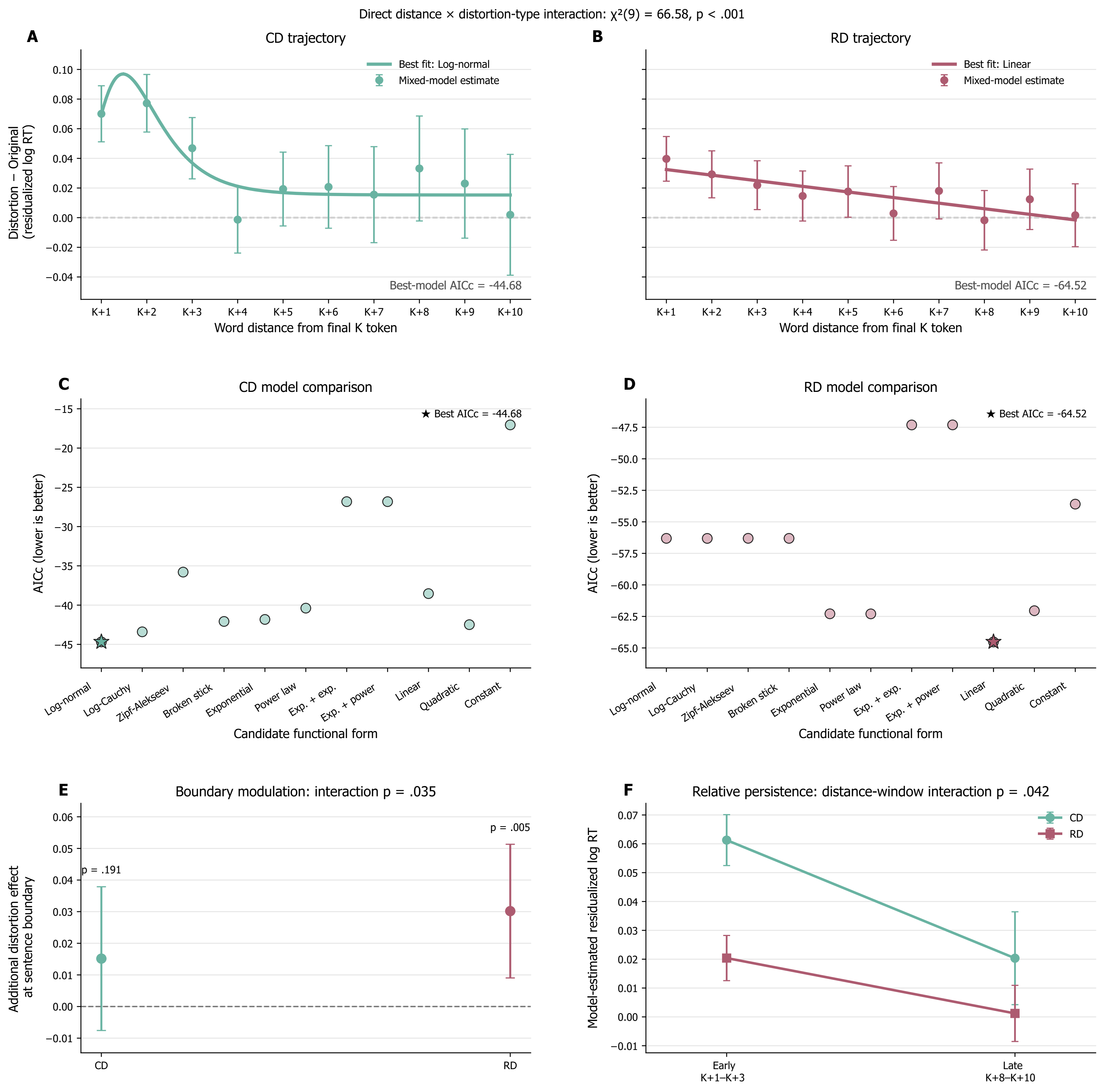}
	\caption{Downstream reading-time dynamics following conceptual and referential distortions. (A–B) Distance-specific effects of conceptual distortion (CD) and referential distortion (RD), relative to matched Original observations, on residualized log reading time from K+1 to K+10. Points show mixed-effects-model estimates with 95\% confidence intervals, and solid lines show the best-fitting functional forms selected by AICc. CD was best described by a log-normal trajectory, whereas RD was best described by a linear trajectory. The trajectories differed significantly across distance, $\chi^2\left(9\right)=66.58, p < 0.001$. (C–D) AICc values for the 11 candidate functional forms fitted to the CD and RD trajectories. Lower values indicate better model support, and stars mark the best-supported model. (E) Additional distortion effect associated with reaching the sentence boundary. Boundary modulation differed significantly between CD and RD ($p = 0.035$). Displayed p values above each estimate indicate the within-distortion boundary effects. (F) Model-estimated distortion effects averaged over an early downstream window (K+1 to K+3) and a late window (K+8 to K+10). The distance-window × distortion-type interaction was significant ($p = 0.042$), indicating that the early-to-late change differed between CD and RD. Error bars indicate 95\% CI. Dashed lines indicate zero effect.}
	\label{FIG:3}
\end{figure}

Both conceptual and referential distortions produced reading-time costs downstream from the manipulated word, but the temporal profiles of these effects differed substantially across distortion types (Figure~\ref{FIG:3}). A direct test of the distance-by-distortion-type interaction across positions K+1 to K+10 was significant, ($\chi^2\left(9\right)=66.58,\ p<0.001$), indicating that the two distortions showed distinct patterns of propagation over subsequent words. A log-normal function best fit the CD effect trajectory, capturing a pronounced early response that peaked between K+1 and K+2 before rapidly declining. By contrast, the RD trajectory was best described by a linear function, with an estimated decrease of 0.0038 log-RT units per additional word. Sentence boundaries further differentiated the two distortions. The RD effect was significantly amplified at sentence boundaries ($\beta=0.030,\ p=0.005$), corresponding to an additional RT cost of approximately 3.06\%; whereas the corresponding modulation for CD was not significant ($\beta=0.015,\ p=0.191$; Figure~\ref{FIG:3}E). Consistent with their different temporal profiles, the change from the early (K+1 to K+3) to late (K+8 to K+10) window also differed significantly between CD and RD ($\beta=0.022,\ p=0.042$; Figure~\ref{FIG:3}F). For human RT, conceptual distortion showed a stronger but more temporally concentrated reading-time response, with an early peak followed by rapid attenuation. Referential distortion produced a weaker and more gradual effect that was selectively amplified at sentence boundaries.

\subsection{Temporal dynamics of conceptual and referential distortion effects in LLM surprisal }\label{results_H}
\begin{figure}
	\centering
	\includegraphics[width=0.87\textwidth]{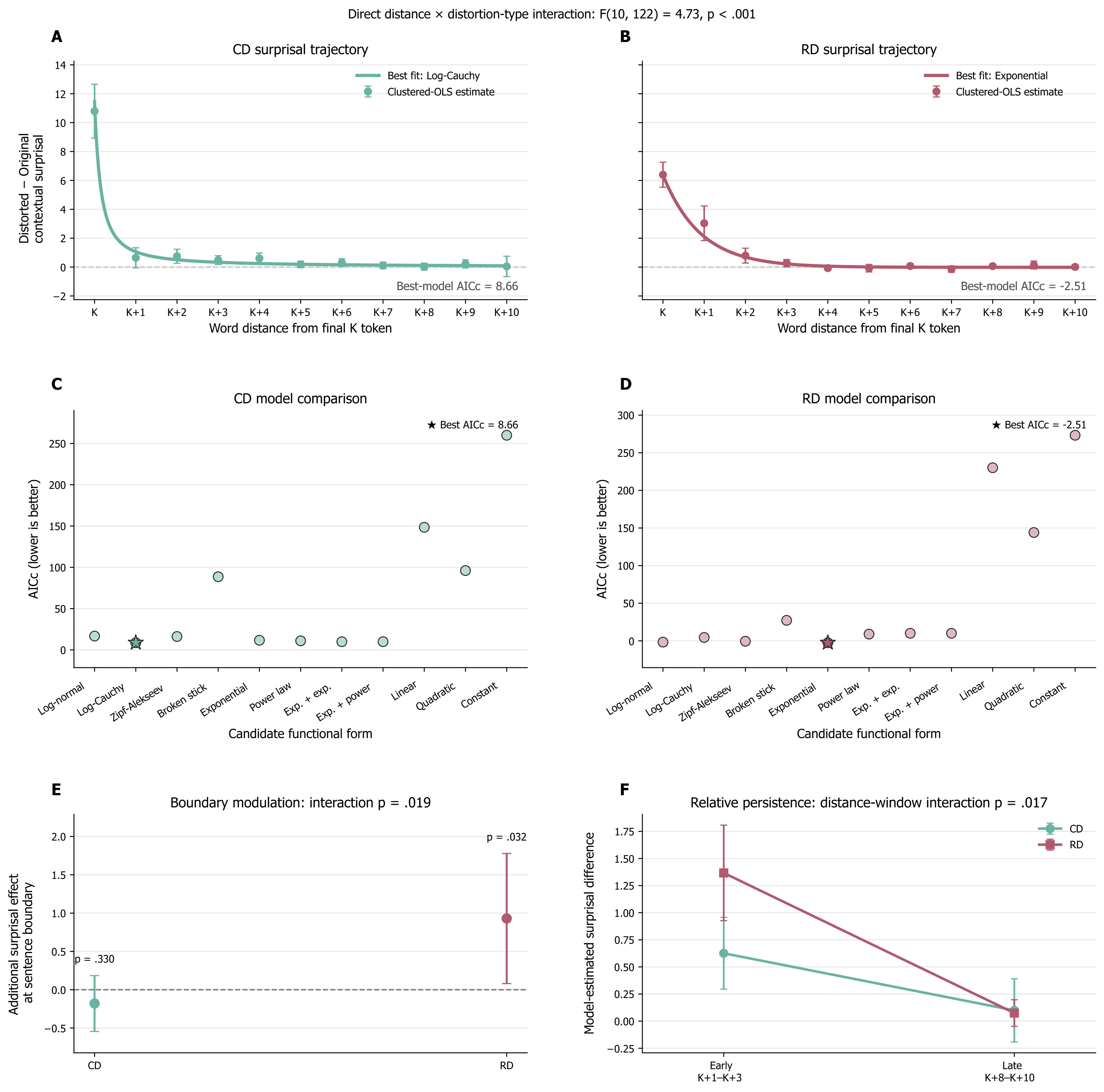}
	\caption{Downstream contextual surprisal dynamics following conceptual and referential distortions. (A–B) Distance-specific effects of conceptual distortion (CD) and referential distortion (RD), relative to matched Original observations, on contextual surprisal from K to K+10. Points show clustered-OLS estimates with 95\% confidence intervals, and solid lines show the best-fitting functional forms selected by AICc. CD was best described by a log-Cauchy trajectory, whereas RD was best described by an exponential trajectory. The trajectories differed significantly across distance, $F(10,122) = 4.73,\ p < 0.001$. (C–D) AICc values for the 11 candidate functional forms fitted to the CD and RD trajectories. Lower values indicate better model support, and stars mark the best-supported model. (E) Additional surprisal effect associated with reaching the sentence boundary. Boundary modulation differed significantly between CD and RD ($p = 0.019$). Displayed p values above each estimate indicate the within-distortion boundary effects. (F) Model-estimated surprisal effects averaged over an early downstream window (K+1 to K+3) and a late window (K+8–K+10). The distance-window × distortion-type interaction was significant ($p = 0.017$), indicating that the early-to-late change differed between CD and RD. Error bars indicate 95\% CI. Dashed lines indicate zero effect.}
	\label{FIG:4}
\end{figure}

Conceptual and referential distortions also produced distinct temporal profiles in contextual surprisal (Figure~\ref{FIG:4}). A direct test of the distance-by-distortion-type interaction across K to K+10 was significant, $F\left(10,122\right)=4.73,p<0.001$, indicating different downstream trajectories for the two distortions. The CD effect was best fit by a log-Cauchy function, capturing a large surprisal increase at K followed by a sharp decline and a small residual effect at subsequent positions. By contrast, the RD trajectory was best described by an exponential function, with a smaller initial effect but a more gradual early decline. Sentence boundaries further differentiated the two distortions. The RD surprisal effect increased significantly at sentence boundaries ($p=0.032$), whereas no significant modulation was observed for CD ($p=0.330$; Figure~\ref{FIG:4}E), yielding a significant distortion-type-by-boundary interaction ($p=0.019$). The change from the early (K+1 to K+3) to late (K+8 to K+10) window also differed significantly between CD and RD ($p=0.017$; Figure~\ref{FIG:4}F). Contextual surprisal therefore showed a strong and sharply localized CD response, whereas RD produced a smaller initial perturbation with greater downstream and boundary sensitivity.

\subsection{Temporal dynamics of conceptual and referential distortion effects in LLM representations}\label{results_cos}
\begin{figure}
	\centering
	\includegraphics[width=0.87\textwidth]{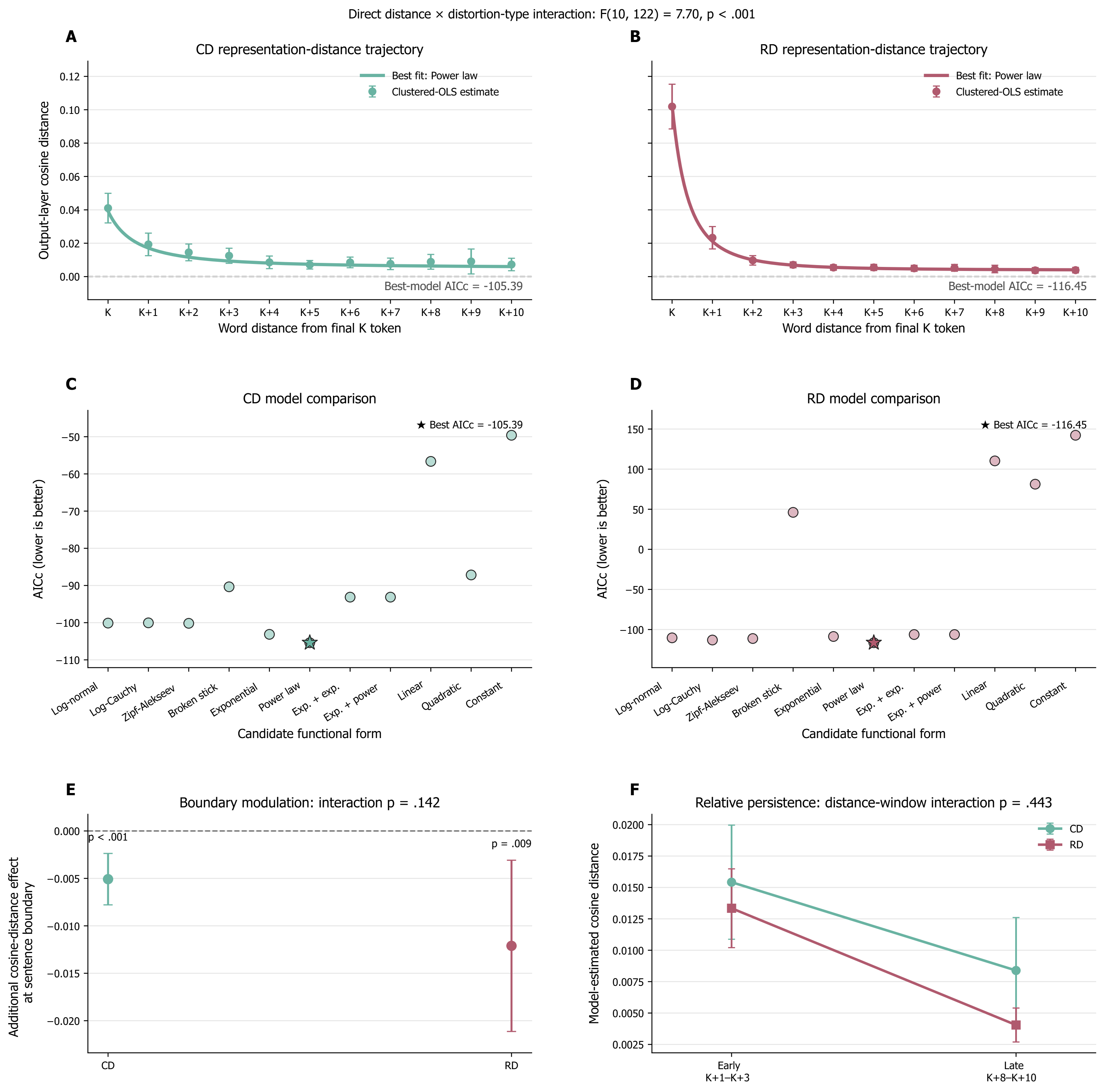}
	\caption{Downstream representational dynamics following conceptual and referential distortions. (A–B) Distance-specific effects of conceptual distortion (CD) and referential distortion (RD), relative to matched Original representations, on output-layer representation distance from K to K+10. Representation distance was quantified as cosine distance between the corresponding final-layer hidden states. Points show clustered-OLS estimates with 95\% confidence intervals, and solid lines show the best-fitting functional forms selected by AICc. Both CD and RD were best described by power-law trajectories. The trajectories differed significantly across distance, $F(10,122)=7.70, p < 0.001$. (C–D) AICc values for the 11 candidate functional forms fitted to the CD and RD trajectories. Lower values indicate better model support, and stars mark the best-supported model. (E) Additional representation-distance effect associated with reaching the sentence boundary. Boundary modulation did not differ significantly between CD and RD ($p = 0.142$). Displayed p values above each estimate indicate the within-distortion boundary effects. (F) Model-estimated representation distances averaged over an early downstream window (K+1 to K+3) and a late window (K+8 to K+10). The distance-window × distortion-type interaction was not significant ($p = 0.443$), indicating no evidence that the early-to-late change differed between CD and RD. Error bars indicate 95\% CI. Dashed lines indicate zero effect.}
	\label{FIG:5}
\end{figure}

Conceptual and referential distortions likewise produced distinct trajectories in output-layer representational distance (Figure~\ref{FIG:5}). The distance-by-distortion-type interaction across K to K+10 was significant, $F(10,122)=7.70,\ p<0.001$. Both trajectories were best described by power-law decay, characterized by a pronounced representational displacement at K followed by rapid attenuation. The initial displacement was substantially larger for RD than for CD, but both effects declined to relatively small residual distances over subsequent words. Sentence boundaries reduced representational distance for both CD ($p<0.001$) and RD ($p=0.009$), with no significant difference in boundary modulation between distortion types ($p=0.142$; Figure~\ref{FIG:5}E). Likewise, the early-to-late change did not differ significantly between CD and RD ($p=0.443$; Figure~\ref{FIG:5}F). Output-layer representations therefore showed a similarly rapidly decaying response to both distortions, distinguished primarily by the substantially larger initial displacement induced by RD. 

\subsection{Replications in Llama-3.2}\label{results_replicate} 
The principal model-based findings generalized to Llama-3.2-3B (Figure S1--S3). Both distortions produced immediate effects at the manipulated word, and contextual surprisal again showed distinct downstream trajectories for CD and RD, with a more concentrated CD response and greater relative persistence for RD. Final-layer representational distance also showed significant distance-by-distortion-type interactions, with both distortions following power-law decay. Sentence boundary effects followed the same general direction as in Qwen3-4B, although the distortion-type interactions did not reach significance. Overall, the replication preserved the principal distance-dependent distinction between CD and RD, while providing weaker statistical evidence for differential boundary modulation.

\section{Discussion}\label{Discussion}
The present study showed that conceptual and referential distortions produced distinct propagation profiles across subsequent linguistic context. In both human RT and LLM surprisal, CD elicited a stronger but more localized effect that declined rapidly, whereas referential distortion produced a weaker, more gradual effect with greater sensitivity to sentence boundaries. Output-layer representations differed from this pattern: referential distortion caused a larger initial displacement, while both trajectories were best described by power-law decay and showed no significant difference in early-to-late persistence. These findings suggest distinguishable processing dynamics between CD and RD, rather than a simple difference in overall disruption magnitude. CD imposes a more temporally concentrated integration cost, whereas RD has more distributed consequences for maintaining discourse-level identity.

This divergence makes theoretical sense. CD changes lexical-conceptual content at a determinate point, concentrating integration cost near the resulting incompatibility. RD, by contrast, disrupts the grammatical relations anchoring an expression to an established discourse referent without making the expression lexically anomalous. Reference, on this account, is not retrieved from an isolated word but established through a grammatical configuration that links an expression to a discourse entity. A related distinction has also been observed in a visual-world eye-tracking paradigm \citep{brocher_dual-process_2018}, where concept pre-activation facilitated lexical access and integration, but referent activation (related to definiteness) shaped the subsequent accessibility of discourse referents. A local referential mismatch may therefore remain temporarily unresolved while this configuration unfolds \citep{greene_pronoun_1992}, distributing its processing cost across subsequent words rather than concentrating it at a single position. 

In this respect, sentence closure completes the current grammatical cycle and initiates a transition to the next one  \citep{arsenijevic_absence_2012}, whereas referential continuity must be maintained across that transition. The boundary increase may therefore reflect the difficulty of carrying forward a referential relation whose grammatical anchoring has been disrupted. This interpretation also fits a broader distinction between conceptual fit, which can be registered relatively locally, and referential resolution, which depends more on subsequent discourse integration \citep{nieuwland_sense_2007, nieuwland_interplay_2008, venhuizen_referential_2025}. A smaller local effect therefore need not imply a weaker disruption overall, but may instead reflect a cost distributed over discourse updating, extending the local-global distinction reported by \cite{soberats_referential_2025} from isolated sentences to a propagation profile in larger discourse. The delayed effect at K+1, instead of K, is not uncommon in SPR studies. In word-by-word SPR, plausibility effects have likewise emerged only after the critical noun \citep{haeuser_how_2022}, while the critical interaction between contextual and real-world knowledge in \cite{jiang_expecting_2025} became apparent only in sentence-final regions. The behavioral consequences of an incompatibility therefore need not coincide with the word that introduces it, but may become measurable as integration continues or a grammatical unit closes. 

The surprisal trajectories provided a computational counterpart to the human RT profiles. In both human RT and LLM surprisal, CD and RD differed not only in magnitude, but also in how their effects propagated through subsequent context. CD directly violated the lexical-conceptual continuation at K, producing a marked predictive penalty; whereas RD could remain syntactically admissible and less unexpected, and exert its effects more gradually as the discourse continued. The disruption concerned the relation established between that expression and a discourse referent, rather than the lexical probability of the expression considered in isolation. For an autoregressive model, such a disruption need not be concentrated in the probability of a single token, but can alter the distribution over subsequent continuations as the discourse is updated. This is consistent with evidence that LLMs condition referential preferences on discourse structure and remain sensitive to intervening referential cues in subsequent pronoun processing \citep{davis_discourse_2020, gautam_robust_2024}. Importantly, downstream words were identical across conditions, so their altered surprisal reflects the different contexts carried forward from the distortion, instead of continuing lexical differences. The stronger sentence-final RD effect further suggests that these consequences were shaped by grammatical structure beyond linear distance alone. Human RT showed the same broad distinction from K+1 onward, consistent with the different temporal alignment of token surprisal and self-paced reading. The correspondence therefore lies in relative magnitude, persistence, and boundary sensitivity, indicating shared sensitivity to the organization of conceptual and referential information, despite the lack of exact temporal identity. 

A different picture emerged at the LLM representational level. Whereas CD produced greater surprisal at K and a larger RT cost from K+1, RD caused a larger initial displacement in output-layer hidden states. This reversal need not be paradoxical. RT reflects the processing cost expressed in readers’ behavior, surprisal reflects the change in the model’s probability distribution, and cosine distance measures the geometric displacement of the hidden state as a whole. The larger initial RD distance should therefore not be interpreted as greater processing difficulty. One possibility is that changing a pronoun or determiner directly reconfigures a bundle of information concerning discourse identity, person or gender, definiteness, and morphosyntactic status. Such a change may strongly alter how the current expression is encoded relative to established discourse entities, even when the expression remains locally probable and produces only a modest predictive penalty. More generally, discourse-level entity information can be encoded in contextualized representations without necessarily producing a corresponding effect in model behavior \citep{davis_discourse_2020, elazar_amnesic_2021}. More broadly, this separation bears on ongoing debates about whether the relational structure learned by LLMs amounts to reference in any stronger sense \citep{mandelkern_language_2024, piantadosi_meaning_2022}. The present study was not designed to resolve that question, but our results show that referential organization leaves a distinct signature in the model’s internal state. RD produced a larger immediate reconfiguration of that state, even when CD carried the larger predictive and behavioral cost. 

Another noteworthy point is that the initial difference did not persist downstream to the same extent as the effects observed in RT and surprisal. CD and RD showed similar power-law decay, while sentence boundaries reduced the representational distance of both CD and RD from the Original condition rather than selectively modulating RD. Neither the boundary nor early-late interaction reliably distinguished the two distortions. Thus, unlike RT and surprisal, output-layer representations provided no evidence for RD-specific persistence. Sentence closure may expose unresolved referential consequences at behavioral and predictive levels while simultaneously promoting representational convergence as shared downstream context accumulates. The larger initial RD distance is therefore better understood as an immediate reconfiguration of discourse-related information than as stronger referential representation or greater processing difficulty.

Taken together, these findings support a dynamic distinction between conceptual and referential meaning. The distinction is not modular: conceptual content helps to descriptively identify discourse entities, whereas reference concerns how linguistic expressions are linked to particular entities and maintained across discourse. These dimensions interact during comprehension, and disruption at one level can affect processing at the other. Their interaction, however, does not imply identical processing dynamics. CD creates a conflict in lexical-conceptual content and produces a stronger, earlier response that declines relatively rapidly over subsequent context. RD disrupts the relation between an expression and its intended discourse referent, with its consequences weaker but more distributed and boundary-sensitive. The contrasting trajectories therefore support a functional distinction between conceptual content and referential organization. In particular, the RD profile is consistent with the view that reference cannot be reduced to lexical content alone, because referents are grammatically anchored and maintained through relations that unfold across discourse. This distinction may also be informative for understanding language impairment in clinical populations, where abnormalities in the use of linguistic context have been reported in patients with schizophrenia \citep{sharpe_selective_2026} and dementia \citep{luzzi_neural_2020}. Methodologically, this distinction would be obscured by analyses restricted to the manipulated word or a predetermined spillover region, which can treat a sharp local peak and a distributed effect as comparable violation costs. Tracing the dynamic trajectory reveals differences in magnitude, persistence, and boundary sensitivity. Propagation is therefore not merely a new way of measuring disruption, but provides the empirical basis for a theoretical account in which conceptual and referential meaning are closely coupled yet distinguishable through downstream consequences.

These findings should be considered with several limitations. The main limitation is that CD and RD were not process-pure. They also differed in lexical category, local salience, grammatical features, lexical frequency, and manipulation rate, so some trajectory differences may reflect these formal properties rather than conceptual–referential status alone. Nevertheless, whole-story perplexity was comparable and lexical surprisal was controlled in the RT analysis. A second issue is that each story contained multiple distortions. Later critical positions may therefore retain effects of earlier perturbations, probably contributing to the non-zero representation distance before K. At the same time, this design allowed propagation to be studied in an accumulating discourse rather than in isolated sentences; single-distortion materials are nevertheless needed to establish the independence of each effect. Generalizability is further limited by the use of 21 fables with manually inserted distortions. The measures also constrain interpretation. SPR cannot determine whether a distortion was detected at K but expressed behaviorally at K+1, and fitting eleven candidate functions to around 11 distance points makes the selected curve families descriptive summaries rather than evidence for specific cognitive mechanisms. Likewise, final-layer cosine distance provides only one view of representational change. Future work should better match local surprisal, lexical category, frequency, and manipulation rate; isolate different referential devices; and use eye-tracking or EEG to resolve detection and integration more precisely. Token-controlled designs and comparisons across model architectures, scales, layers, and distance metrics will be necessary to determine how broadly the present distinction generalizes.

\section*{Data and code availability }
De-identified reading-time data are publicly available in Mendeley Data at \url{https://doi.org/10.17632/pb8mvz4tby.1}. Scripts for data preprocessing and analysis are available at \url{https://github.com/RuiHe1999/SPR_distortion}.

\section*{Acknowledgment}
This study was supported by European Research Council (ERC-2023-SyG, 101118756). Views and opinions expressed are however those of the authors only and do not necessarily reflect those of the European Union or the Agency. Neither the European Union nor the granting authority can be held responsible for them. 

\bibliographystyle{elsarticle-harv} 
\bibliography{citations}

\clearpage
\appendix
\renewcommand{\thefigure}{S\arabic{figure}}
\setcounter{figure}{0}

\section{Example excerpts for conceptual and referential distortions}

\subsection*{Original:}
An ant went to the riverbank to quench its thirst but was swept away by the rushing \textbf{current}, nearly drowning. A dove, perched on a tree hanging over the water, noticed \textbf{the} ant's struggle… 

\subsection*{Conceptual distortion:}
An ant went to the riverbank to quench its thirst but was swept away by the rushing \textbf{mountain}, nearly drowning. A dove, perched on a tree hanging over the water, noticed the ant's struggle…

\subsection*{Referential distortion:}
An ant went to the riverbank to quench its thirst but was swept away by the rushing current, nearly drowning. A dove, perched on a tree hanging over the water, noticed \textbf{an} ant's struggle… 

\section{Curve fitting and model comparison}
For each distortion type, the distance-specific effects estimated from the mixed-effects models were fitted with 11 candidate functions. Let $\mathbf{y}=\left(y_1,\ldots,y_n\right)^\top$ denote the vector of estimated distortion effects across the $n$ supported word distances, and let $\Sigma$ denote their estimated covariance matrix. For a candidate function $f(x;\boldsymbol{\theta})$, the corresponding vector of fitted values was
\begin{equation*}
\mathbf{f}_{\boldsymbol{\theta}}=\left(f(x_1;\boldsymbol{\theta}),\ldots,f(x_n;\boldsymbol{\theta})\right)^\top.
\end{equation*}

Let $x$ denote word distance from the final distorted token ($x=1,\ldots,10$), and let $f\left(x\right)$ denote the predicted distortion effect at distance x. The candidate functions were defined as follows:
\begin{enumerate}
    \item Constant: $f\left(x\right)=b$
	\item Linear: $f\left(x\right)=b+sx$
	\item Quadratic: $f\left(x\right)=b+sx+qx^2$
	\item Exponential: $f\left(x\right)=b+Ae^{-rx}$
	\item Power law: $f\left(x\right)=b+A\left(x+1\right)^{-\alpha}$
	\item Log-normal: $f\left(x\right)=b+A\exp{\left[-\frac{1}{2}\left(\frac{\log{x}-\mu}{\sigma}\right)^2\right]}$
	\item Log-Cauchy: $f\left(x\right)=b+A\frac{\sigma^2}{\sigma^2+\left(\log{x}-\mu\right)^2}$
	\item Zipf-Alekseev: $f\left(x\right)=b+A\exp{\left[-\alpha\log{\left(x+1\right)}-\beta\{\log{\left(x+1\right)}\}^2\right]}$
	\item Broken stick: $f\left(x\right)=b+s_1x+\left(s_2-s_1\right)\max{\left(x-\kappa,0\right)}$
	\item Double exponential: $f\left(x\right)=b+A\left[we^{-r_1x}+\left(1-w\right)e^{-r_2x}\right]$
	\item Exponential + power: $f\left(x\right)=b+A\left[w\left(x+1\right)^{-\alpha}+\left(1-w\right)e^{-rx}\right]$
\end{enumerate}

For the reading-time trajectories, which began at $x=1$ (i.e., K+1), the log-normal and log-Cauchy functions were defined on $\log{\left(x\right)}$. For the language-model trajectories, which included the perturbation position itself ($x=0$; i.e., K), these two functions were evaluated on $\log{\left(x+1\right)}$ to ensure that the functions were defined at $x=0$.

Here, $b$ denotes the baseline level, $A$ the amplitude of the distance-dependent component, $s$ and $q$ the linear and quadratic coefficients, respectively, $r$, $r_1$, and $r_2$ exponential decay rates, $\alpha$ and $\beta$ power or shape parameters, $\mu$ and $\sigma$ the location and scale parameters of the log-domain functions, $s_1$ and $s_2$ the slopes before and after the broken-stick knot $\kappa$, and $w$ a mixture weight constrained between 0 and 1.

The constant, linear, and quadratic models were left unconstrained. For the exponential and power-law models, decay or power parameters were constrained to be positive. The scale parameter $\sigma$ in the log-normal and log-Cauchy models was constrained to $0.05\le\sigma\le5$, while their location parameter $\mu$ was allowed to vary over a range extending three log-distance units beyond the observed distance range. For the Zipf–Alekseev model, $\alpha$ and $\beta$ were constrained to the interval $\left[0,10\right]$. The knot $\kappa$ of the broken-stick model was restricted to lie between the second and penultimate observed distances. For the double-exponential and exponential-plus-power models, decay or power parameters were constrained to be positive and mixture weights to $0 \leq w \leq 1$.

Parameters were estimated by generalized least squares (GLS), minimizing
\begin{equation*}
\chi_{\mathrm{GLS}}^2=
\left(\mathbf{y}-\mathbf{f}_{\theta}\right)^\top
\Sigma^{-1}
\left(\mathbf{y}-\mathbf{f}_{\theta}\right).
\end{equation*}

This procedure incorporated both the uncertainty of individual distance-specific estimates and the covariance among estimates obtained from the same statistical model. Before fitting, the covariance matrix was symmetrized and, when necessary, regularized to ensure numerical positive definiteness by eigenvalue decomposition, with very small or negative eigenvalues replaced by a small positive floor.

Nonlinear candidate functions were fitted using bounded nonlinear least squares with the full covariance matrix supplied as the weighting matrix. To reduce sensitivity to parameter initialization and local optima, nonlinear models were fitted from multiple starting values, and the solution yielding the smallest $\chi_{\mathrm{GLS}}^2$ was retained. 

Assuming multivariate Gaussian uncertainty in the distance-specific estimates, model fit was quantified as
\begin{equation*}
-2\log{L}=n\log{\left(2\pi\right)}+\log{\left|\Sigma\right|}+\chi_{\mathrm{GLS}}^2.
\end{equation*}

For a candidate model with k fitted parameters, the Akaike information criterion (AIC) was
\begin{equation*}
\mathrm{AIC}=-2\log{L}+2k,
\end{equation*}
and the small-sample corrected Akaike information criterion (AICc) was
\begin{equation*}
\mathrm{AICc}=\mathrm{AIC}+\frac{2k\left(k+1\right)}{n-k-1}.
\end{equation*}

AICc was used as the primary model-selection criterion because the number of available distance points was small relative to the number of free parameters in some candidate functions. Models for which $n\leq k+1$ were therefore not considered estimable under AICc. The model with the lowest AICc was considered the best-supported functional form. 

For the selected functional form of each reading-time trajectory, uncertainty in derived trajectory characteristics was estimated using a parametric bootstrap. Let $\hat{\mathbf{y}}$ denote the vector of distance-specific effects and $\hat{\Sigma}$ its estimated covariance matrix. For each of 1,000 bootstrap iterations, a new effect trajectory was sampled from
\begin{equation*}
\mathbf{y}^{(b)}\sim\mathcal{N}\left(\hat{\mathbf{y}},\hat{\Sigma}\right), \qquad b=1,\ldots,1000.
\end{equation*}

The selected functional form was refitted to each sampled trajectory using the same covariance-weighted fitting procedure and parameter constraints. Bootstrap iterations for which fitting failed or yielded non-finite derived quantities were excluded. A fixed random seed of 42 was used for reproducibility.

For a selected log-normal trajectory, the peak distance was defined as $x_{\mathrm{peak}}=\exp{\left(\mu\right)}$, with corresponding peak effect $f\left(x_{\mathrm{peak}}\right)=b+A$. The lower and upper distances at which the distance-dependent component reached half of its maximum were:
\begin{equation*}
x_{\mathrm{half},-}=\exp{\left(\mu-\sigma\sqrt{2\log{2}}\right)}, \qquad x_{\mathrm{half},+}=\exp{\left(\mu+\sigma\sqrt{2\log{2}}\right)},
\end{equation*}
respectively. For a selected linear trajectory, the slope (s) quantified the change in the distortion effect per additional word. Predicted effects at the first and final analyzed distances and their total change were also derived. 95\% confidence intervals for derived trajectory characteristics were obtained from their bootstrap distributions.

\section{Results with Llama 3.2}

\begin{figure}[h]
    \centering
    \includegraphics[width=1\linewidth]{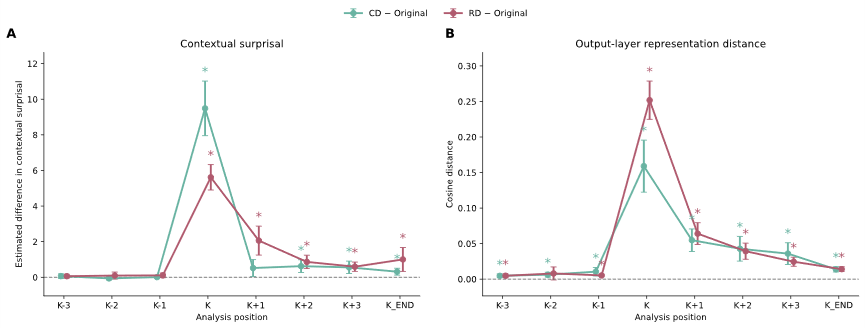}
    \caption{Local effects of surprisal and representational distance with Llama 3.2 3B.}
    \label{fig:s1}
\end{figure}

\begin{figure}[h]
    \centering
    \includegraphics[width=1\linewidth]{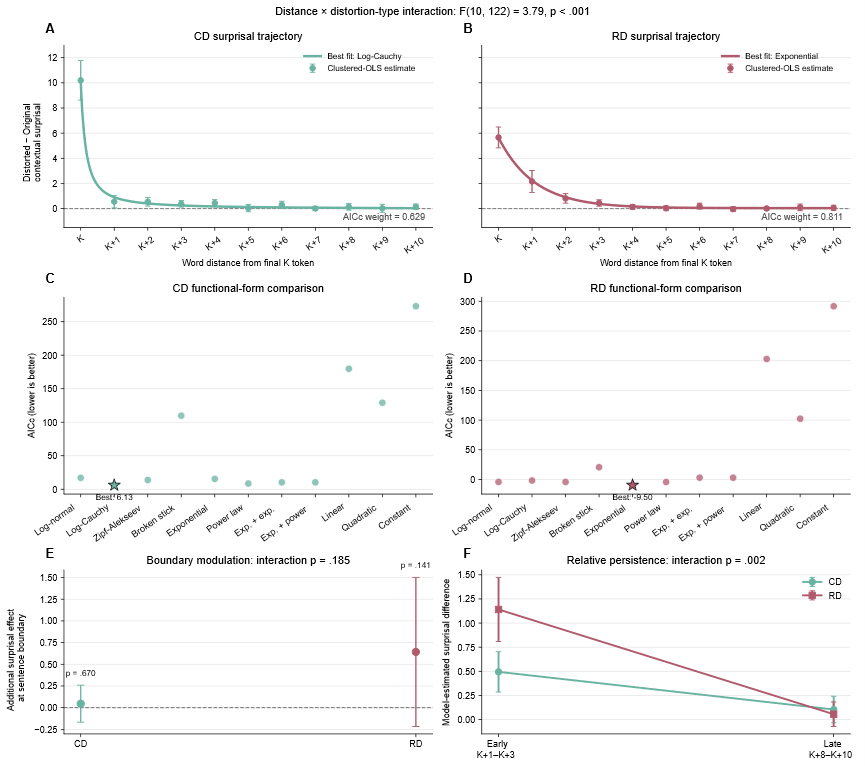}
    \caption{Effect decay of Llama 3.2 surprisal.}
    \label{fig:s2}
\end{figure}

\begin{figure}[h]
    \centering
    \includegraphics[width=1\linewidth]{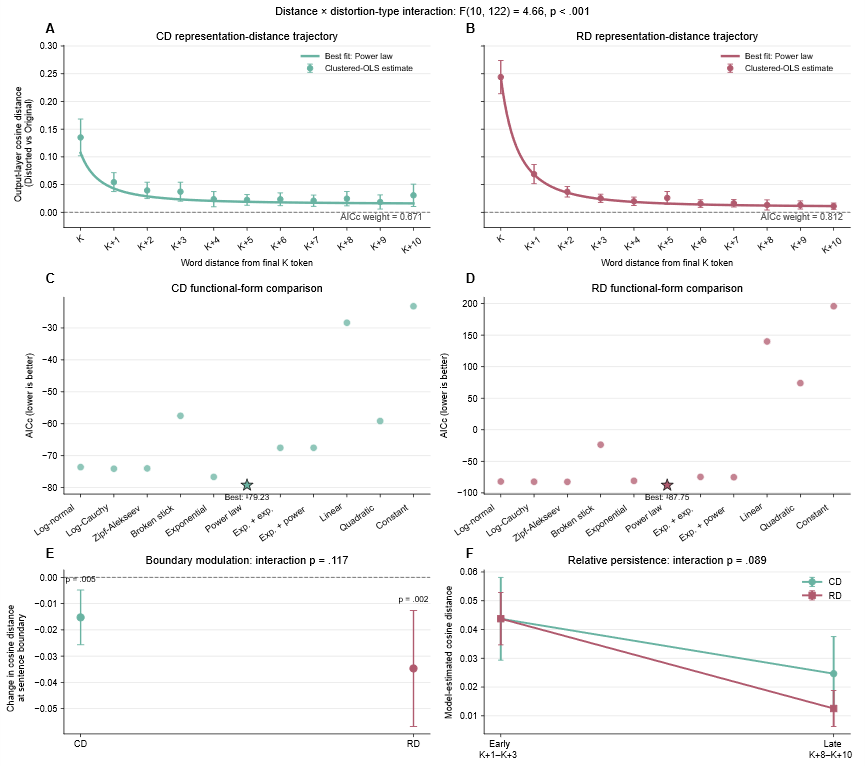}
    \caption{Effect decay of Llama 3.2 representational distance.}
    \label{fig:s3}
\end{figure}

\end{document}